\documentclass[11pt,a4paper]{article}
\usepackage[hyperref]{acl2018}
\usepackage{times}
\usepackage{latexsym}

\usepackage{url}

\usepackage{url}
\usepackage{color}
\usepackage{amsmath}
\usepackage{amssymb}
\usepackage{graphicx}
\usepackage{multirow}
\usepackage[linesnumbered,ruled,vlined]{algorithm2e}
\usepackage{algpseudocode}
\usepackage{todonotes}
\usepackage{booktabs}
\usepackage{caption}
\DeclareCaptionFont{10pt}{\fontsize{10pt}{12pt}\selectfont}
\usepackage[flushleft]{threeparttable}
\usepackage{textcomp}

\newcommand\QNfinal[1]{#1}
\newcommand\QN[1]{\textrm{#1}}

\newcommand\argmax[2]{\textrm{arg}\max_{#1}{#2}}
\newcommand{\event}[1]{\textit{\textbf{#1}}}
\newcommand{\idxevent}[2]{\event{e#1:#2}}
\newcommand{\rel}[1]{\textit{#1}}
\newcommand{\mc}[1]{\mathcal{#1}}

\newcommand{\ignore}[1]{}

\newcommand{\CAEVO}{chambers2014dense}
\newcommand{\ClearTK}{bethard2013cleartk}
\newcommand{\CATENA}{mirza2016catena}
\newcounter{exctr}
\newcommand{\addexctr}{\refstepcounter{exctr}\theexctr}

\aclfinalcopy 
\def\aclpaperid{***} 

\title{\vspace*{-0.5in}{\small \hfill ACL'18}\\
	\vspace*{0.25in} Joint Reasoning for Temporal and Causal Relations}

\author{Qiang Ning,$^1$ Zhili Feng,$^2$ Hao Wu,$^3$ Dan Roth$^{1,3}$ \\
	Department of Computer Science\\
	$^1$University of Illinois at Urbana-Champaign, Urbana, IL 61801, USA\\
	$^2$University of Wisconsin-Madison, Madison, WI 53706, USA\\
	$^3$University of Pennsylvania, Philadelphia, PA 19104, USA\\
	{\tt \small qning2@illinois.edu,~zfeng49@cs.wisc.edu,~\{haowu4,danroth\}@seas.upenn.edu}}

\date{}

\begin{document}
\maketitle

\begin{abstract}
	Understanding temporal and causal relations between events is a fundamental natural language understanding task. 
	Because a cause must occur earlier than its effect, temporal and causal relations are closely related and one relation often dictates the value of the other. 
	However, limited attention has been paid to studying these two relations jointly. 
	This paper presents a joint inference framework for them using constrained conditional models (CCMs).
	Specifically, we formulate the joint problem as an integer linear programming (ILP) problem, enforcing constraints that are inherent in the nature of time and causality.
	We show that the joint inference framework results in statistically significant improvement in the extraction of both temporal and causal relations from text.\footnote{\QNfinal{The dataset and code used in this paper are available at \url{http://cogcomp.org/page/publication_view/835}}}
\end{abstract}
\section{Introduction}
Understanding events is an important component of natural language understanding.
An essential step in this process is identifying relations between events, which are needed in order to support applications such as story completion, summarization, and timeline construction.

Among the many relation types that could exist between events, this paper focuses on the joint extraction of temporal and causal relations.
It is well known that temporal and causal relations interact with each other and in many cases, the decision of one relation is made primarily based on evidence from the other.
In Example~\ref{ex:prior knowledge 1}, identifying the temporal relation between \idxevent{1}{died} and \idxevent{2}{exploded} is in fact a very hard case: There are no explicit temporal markers (e.g., ``before'', ``after'', or ``since''); the events are in separate sentences so their syntactic connection is weak; although the occurrence time of  \idxevent{2}{exploded} is given (i.e., Friday) in text, it is not given for  \idxevent{1}{died}. However, given the causal relation, \idxevent{2}{exploded} caused \idxevent{1}{died},it is clear that \idxevent{2}{exploded} happened before \idxevent{1}{died}. The temporal relation is dictated by the causal relation.

\begin{table}[h!]
	\centering\small
	\begin{tabular}{|p{7cm}|}
		\hline
		\textbf{Ex~\addexctr\label{ex:prior knowledge 1}: Temporal relation dictated by causal relation.}\\
		\hline
		More than 10 people \event{(e1:died)} on their way to the nearest hospital, police said. A suicide car bomb \event{(e2:exploded)} on Friday in the middle of a group of men playing volleyball in northwest Pakistan.\\
		\hline
		{\em Since \idxevent{2}{exploded} is the reason of \idxevent{1}{died}, the temporal relation is thus e2 being before e1.}\\
		\hline\hline
		\textbf{Ex~\addexctr\label{ex:prior knowledge 2}: Causal relation dictated by temporal relation.}\\
		\hline
		Mir-Hossein Moussavi \event{(e3:raged)} after government's efforts to \event{(e4:stifle)} protesters.\\
		\hline
		{\em Since \idxevent{3}{raged} is temporally after \idxevent{4}{stifle}, e4 should be the cause of e3.}\\
		\hline
	\end{tabular}
\end{table}

On the other hand, causal relation extraction can also benefit from knowing temporal relations. In Example~\ref{ex:prior knowledge 2}, it is unclear whether the government stifled people because people raged, or people raged because the government stifled people\QNfinal{: both situations are logically reasonable}. However, if we account for the temporal relation (that is, \idxevent{4}{stifle} happened before \idxevent{3}{raged}), it is clear that \idxevent{4}{stifle} is the cause and \idxevent{3}{raged} is the effect. In this case, the causal relation is dictated by the temporal relation.

The first contribution of this work is proposing a joint framework for \textbf{T}emporal and \textbf{C}ausal \textbf{R}easoning (TCR), inspired by these examples.
Assuming the availability of a temporal extraction system and a causal extraction system, the proposed joint  framework combines these two using a constrained conditional model (CCM)~\cite{ChangRaRo12} framework, with an integer linear programming (ILP) objective~\cite{RothYi04} that enforces declarative constraints during the inference phase. 
Specifically, these constraints include: (1) A cause must temporally precede its effect. (2) Symmetry constraints, i.e., when 
a pair of events, $(A,B)$, has a temporal relation $r$ (e.g., \rel{before}), then $(B,A)$ must have the reverse relation of $r$ (e.g., \rel{after}). (3) Transitivity constraints, i.e., the relation between $(A,C)$ must be temporally consistent with the relation derived from $(A,B)$ and $(B,C)$.
These constraints originate from the one-dimensional nature of time and the physical nature of causality and build connections between temporal and causal relations, making CCM a natural choice for this problem.
As far as we know, very limited work has been done in joint extraction of both relations. Formulating the joint problem in the CCM framework is novel and thus the first contribution of this work.

A key obstacle in jointly studying temporal and causal relations lies in the absence of jointly annotated data. 
The second contribution of this work is the development of such a jointly annotated dataset which we did by augmenting the EventCausality dataset \cite{DoChRo11} with dense temporal annotations. This dataset allows us to show statistically significant improvements on both relations via the proposed joint framework. 

\QN{This paper also presents an empirical result of improving the temporal extraction component.
Specifically, we incorporate explicit time expressions present in the text and high-precision knowledge-based rules into the ILP objective. 
These sources of information have been successfully adopted by existing methods \cite{\CAEVO,\CATENA}, but were never used within a global ILP-based inference method.
Results on TimeBank-Dense \cite{cassidy2014annotation}, a benchmark dataset with temporal relations only, show that these modifications can also be helpful within ILP-based methods.
}

\section{Related Work}
Temporal and causal relations can both be represented by directed acyclic graphs, where the nodes are events and the edges are labeled with either {\em before, after}, etc. (in temporal graphs), or {\em causes} and {\em caused by} (in causal graphs).
Existing work on {\em temporal} relation extraction was initiated by \cite{mani2006machine,chambers2007classifying,bethard2007timelines,verhagen2008temporal}, which formulated the problem as that of learning a classification model for determining the label of each edge locally (i.e., {\em local} methods). 
A disadvantage of these early methods is that the resulting graph may break the symmetric and transitive constraints. 
There are conceptually two ways to enforce such graph constraints (i.e., {\em global} reasoning). 
CAEVO \cite{chambers2014dense} grows the temporal graph in a multi-sieve manner, where predictions are added sieve-by-sieve. A graph closure operation had to be performed after each sieve to enforce constraints.
\QN{This is solving the global inference problem greedily.}
A second way is to perform exact inference via ILP and the symmetry and transitivity requirements can be enforced as ILP constraints~\cite{BDLB06,ChambersJu08,DenisMu11,DoLuRo12,NingFeRo17}.

We adopt the ILP approach in the temporal component of this work for two reasons. First, as we show later, it is straightforward to build a joint framework with both temporal and causal relations as an extension of it.
Second, the relation between a pair of events is often determined by the relations among other events.
In Ex~\ref{ex:ilp}, if a system is unaware of $(e5,e6)$=\rel{simultaneously} when trying to make a decision for $(e5,e7)$, it is likely to predict that $e5$ is \rel{before} $e7$\footnote{Consider the case that ``The FAA \idxevent{5}{announced}\dots it \idxevent{7}{said} it would\dots''. Even humans may predict that $e5$ is \rel{before} $e7$.}; but, in fact, $(e5,e7)$=\rel{after} given the existence of $e6$. 
Using global considerations is thus beneficial in this context not only for generating globally consistent temporal graphs, but also for making more reliable pairwise decisions.

\begin{table}
	\centering\small
	\begin{tabular}{|p{7cm}|}
		\hline
		\textbf{Ex \addexctr\label{ex:ilp}: Global considerations are needed when making local decisions.}\\
		\hline
		The FAA on Friday \event{(e5:announced)} it will close 149 regional airport control towers because of forced spending cuts. Before Friday's \event{(e6:announcement)}, it \event{(e7:said)} it would consider keeping a tower open if the airport convinces the agency it is in the "national interest" to do so.\\
		\hline
	\end{tabular}
\end{table}

Prior work on {\em causal} relations in {\em natural language text} was relatively sparse. Many causal extraction work in other domains assumes the existence of ground truth timestamps (e.g., \cite{SunXiLiYaZhCh07,GuYeSw16}), but gold timestamps rarely exist in natural language text.
In NLP, people have focused on causal relation identification using lexical features or discourse relations.
For example, based on a set of explicit causal discourse markers (e.g., ``because'', ``due to'', and ``as a result''), \citet{hidey2016identifying} built parallel Wikipedia articles and constructed an open set of implicit markers called AltLex. A classifier was then applied to identify causality.
\citet{dunietz2017automatically} used the concept of construction grammar to tag causally related clauses or phrases.
\citet{DoChRo11} considered global statistics over a large corpora, the cause-effect association (CEA) scores, and combined it with discourse relations using ILP to identify causal relations. These work only focused on the causality task and did not address the temporal aspect.

However, as illustrated by Examples~\ref{ex:prior knowledge 1}-\ref{ex:prior knowledge 2}, temporal and causal relations are closely related,
\QNfinal{as assumed by many existing works \cite{BethardMa08,RinkBeHa10}}.
Here we argue that being able to capture both aspects in a joint framework provides a more complete understanding of events in natural language documents. 
Researchers have started paying attention to this direction recently. 
For example, \citet{CaTeRs} proposed an annotation framework, CaTeRs, which captured both temporal and causal aspects of event relations in common sense stories. 
CATENA \cite{mirza2016catena} extended the multi-sieve framework of CAEVO to extracting both temporal and causal relations and exploited their interaction through post-editing temporal relations based on causal predictions.
In this paper, we 
push this idea forward and tackle the problem in a joint and more principled way, as shown next.
\section{Temporal and Causal Reasoning}
\label{sec:proposed}

\ignore{
Unlike local methods which only make one pairwise decision at a time, joint inference has the unique capability to see farther and consider ``globally''.
}

In this section, we explain the proposed joint inference framework, \textbf{T}emporal and \textbf{C}ausal \textbf{R}easoning (TCR).
To start with, we focus on introducing the temporal component, and clarify how to design the transitivity constraints and how to enforce other readily available prior knowledge to improve its performance.
With this temporal component already explained, we further incorporate causal relations and complete the TCR joint inference framework.
Finally, we transform the joint problem into an ILP so that it can be solved using off-the-shelf packages.

\subsection{Temporal Component}

Let $\mc{R}_T$ be the label set of temporal relations and $\mc{E}$ and $\mc{T}$ be the set of all events and the set of all time expressions (a.k.a. timex) in a document. For notation convenience, we use $\mc{EE}$ to represent the set of all event-event pairs; then $\mc{ET}$ and $\mc{TT}$ have obvious definitions.
Given a pair in $\mc{EE}$ or $\mc{ET}$, assume for now that we have corresponding classifiers producing confidence scores for every temporal relation in $\mc{R}_T$. 
Let them be $s^{ee}(\cdot)$ and $s^{et}(\cdot)$, respectively.
Then the inference formulation for all the temporal relations within this document is:
\begin{equation}\small
\label{eq:temp formulation}
\hat{Y} = \argmax{Y\in\mc{Y}}{}\sum_{i\in\mc{EE}}{}{s^{ee}\{i\mapsto Y_{i}\}}
+\sum_{j\in\mc{ET}}{s^{et}\{j\mapsto Y_{j}\}}
\end{equation}
where $Y_k\in\mc{R}_T$ is the temporal label of pair $k\in\mc{MM}$ (Let $\mc{M}=\mc{E}\cup\mc{T}$ be the set of all temporal nodes), ``$k\mapsto Y_k$" represents the case where the label of pair $k$ is predicted to be $Y_k$, $Y$ is a vectorization of all these $Y_k$'s in one document, and $\mc{Y}$ is the constrained space that $Y$ lies in.

We do not include the scores for $\mc{TT}$ because the temporal relationship between timexes can be trivially determined using the  normalized dates of these timexes,  as was done in \cite{DoLuRo12,\CAEVO,\CATENA}. We impose these relations via equality constraints denoted as $\mc{Y}_0$.
In addition, we add symmetry and transitivity constraints dictated by the nature of time (denoted by $\mc{Y}_1$), and other prior knowledge derived from linguistic rules (denoted by $\mc{Y}_2$), which will be explained subsequently.
Finally, we set $\mc{Y}=\cap_{i=0}^2\mc{Y}_i$ in Eq.~\eqref{eq:temp formulation}.
 
\textbf{Transitivity Constraints.}
Let the dimension of $Y$ be $n$. Then a standard way to construct the symmetry and transitivity constraints is shown in \cite{BDLB06,ChambersJu08,DenisMu11,DoLuRo12,NingFeRo17}
\begingroup\makeatletter\def\f@size{10}\check@mathfonts
\def\maketag@@@#1{\hbox{\m@th\large\normalfont#1}}%
\begin{align*}\small
\mc{Y}_1=\left\{Y\in \mc{R}_T^n\lvert \forall m_{1,2,3}\in\mc{M}, Y_{(m_1,m_2)}=\bar{Y}_{(m_2,m_1)},\right.\\
\left.Y_{(m_1,m_3)}\in \textrm{Trans}(Y_{(m_1,m_2)}, Y_{(m_2,m_3)})\right\}
\end{align*}\endgroup
where the bar sign is used to represent the reverse relation hereafter, and $\textrm{Trans}(r_1,r_2)$ is a set comprised of all the temporal relations from $\mc{R}_T$ that do not conflict with $r_1$ and $r_2$.

\begin{figure}[htbp!]
	\centering
	\includegraphics[width=0.4\textwidth]{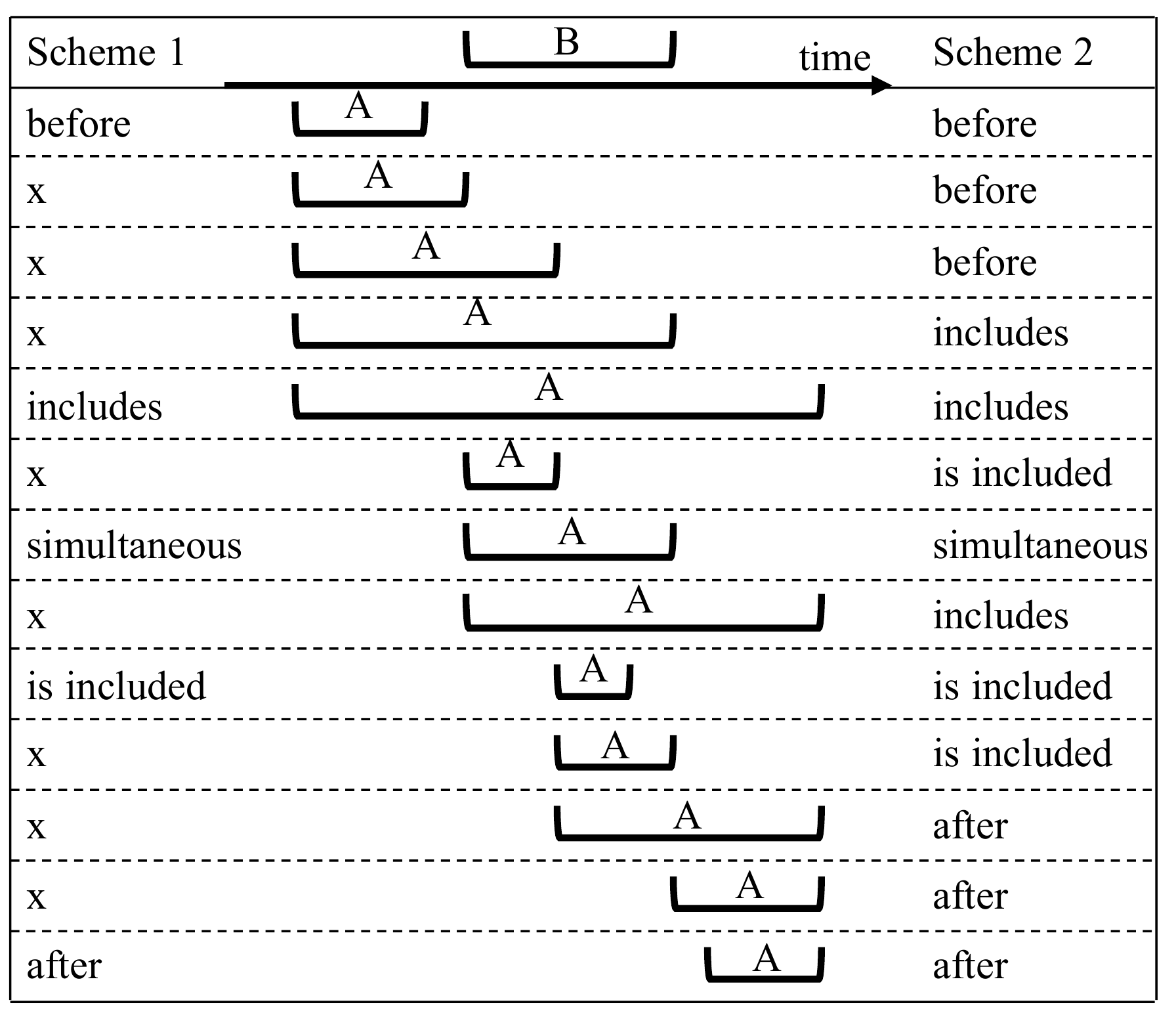}
	\caption{\textbf{Two possible interpretations to the label set of $\mathcal{R}_T=\{b,a,i,ii,s,v\}$} for the temporal relations between (A, B). ``x" means that the label is ignored. Brackets represent time intervals along the time axis. Scheme 2 is adopted consistently in this work.}
	\label{fig:reduce}
\end{figure}

The construction of $\textrm{Trans}(r_1,r_2)$ necessitates a clearer definition of $\mc{R}_T$, the importance of which is often overlooked by existing methods. 
Existing approaches all followed the interval representation of events \cite{allen1984towards}, which yields 13 temporal relations (denoted by $\tilde{\mc{R}}_T$ here) as shown in Fig.~\ref{fig:reduce}. 
Most systems used a reduced set, for example, \{\rel{before}, \rel{after}, \rel{includes}, \rel{is included}, \rel{simultaneously}, \rel{vague}\}. For notation convenience, we denote them $\mc{R}_T=\{b, a, i, ii, s, v\}$.
Using a reduced set is more convenient in data annotation and leads to better performance in practice. 

However, there has been limited discussion in the literature on how to interpret the reduced relation types.
For example, is the ``\rel{before}" in $\mc{R}_T$ exactly the same as the ``\rel{before}" in the original set ($\tilde{\mc{R}}_T$) (as shown on the left-hand-side of Fig.~\ref{fig:reduce}), or is it a combination of multiple relations in $\tilde{\mc{R}}_T$ (the right-hand-side of Fig.~\ref{fig:reduce})? We compare two reduction schemes in Fig.~\ref{fig:reduce}, where scheme 1 ignores low frequency labels directly and scheme 2 absorbs low frequency ones into their temporally closest labels.
The two schemes barely have differences when a system only looks at a single pair of mentions at a time (this might explain the lack of discussion over this issue in the literature), but they lead to different Trans$(r_1,r_2)$ sets and this difference can be magnified when the problem is solved jointly and when the label distribution changes across domains. To completely cover the 13 relations, we adopt scheme 2 in this work.

The resulting transitivity relations are shown in Table~\ref{tab:trans}. 
The top part of Table~\ref{tab:trans} is a compact representation of three generic rules; for instance, Line 1 means that 
the labels themselves are transitive. Note that
\QNfinal{during human annotation, if an annotator looks at a pair of events and decides that multiple well-defined relations can exist, he/she labels it \rel{vague}; also, when aggregating the labels from multiple annotators, a label will be changed to \rel{vague} if the annotators disagree with each other. In either case, \rel{vague} is chosen to be the label when a single well-defined relation cannot be uniquely determined by the contextual information.
This explains why a \rel{vague} relation (v) is always added in Table~\ref{tab:trans} if more than one label in Trans($r_1,r_2$) is possible.
As for Lines 6, 9-11 in Table 1 (where \rel{vague} appears in Column $r_2$), Column Trans($r_1$,$r_2$) was designed in such a way that $r_2$ cannot be uniquely determined through $r_1$ and Trans($r_1$,$r_2$). For instance, $r_1$ is \rel{after} on Line 9, if we further put \rel{before} into Trans($r_1$,$r_2$), then $r_2$ would be uniquely determined to be \rel{before}, conflicting with $r_2$ being \rel{vague}, so \rel{before} should not be in Trans($r_1$,$r_2$).}

\begin{table}
\centering
\small
\begin{tabular}{l|c|c|c}
\hline
No. 	&	$r_1$		&	$r_2$		&	Trans($r_1$, $r_2$)\\
\hline
1	&	$r$	&	$r$	&	$r$\\
2	&	$r$	&	\textbf{s}	&	$r$\\
3	&	$r_1$	&	$r_2$	&	$\overline{\textrm{Trans}(\bar{r}_2,\bar{r}_1)}$\\
\hline\hline
4	&	\textbf{b}	&	\textbf{i}	&	\textbf{b}, \textbf{i}, \textbf{v}\\
5	&	\textbf{b}	&	\textbf{ii}	&	\textbf{b}, \textbf{ii}, \textbf{v}\\
6	&	\textbf{b}	&	\textbf{v}	&	\textbf{b}, \textbf{i}, \textbf{ii}, \textbf{v}\\
7	&	\textbf{a}	&	\textbf{i}	&	\textbf{a}, \textbf{i}, \textbf{v}\\
8	&	\textbf{a}	&	\textbf{ii}	&	\textbf{a}, \textbf{ii}, \textbf{v}\\
9	&	\textbf{a}	&	\textbf{v}	&	\textbf{a}, \textbf{i}, \textbf{ii} ,\textbf{v}\\
10	&	\textbf{i}	&	\textbf{v}	&	\textbf{b}, \textbf{a}, \textbf{i}, \textbf{v}\\
11	&	\textbf{ii}	&	\textbf{v}	&	\textbf{b}, \textbf{a}, \textbf{ii}, \textbf{v}\\
\hline
\end{tabular}
\caption{\textbf{Transitivity relations} based on the label set reduction scheme 2 in Fig.~\ref{fig:reduce}.  If $(m_1,m_2)\mapsto r_1$ and $(m_2,m_3)\mapsto r_2$, then the relation of $(m_1,m_3)$ must be chosen from Trans$(r_1,r_2)$, $\forall m_1$, $m_2$, $m_3\in\mc{M}$. The top part of the table uses $r$ to represent generic rules compactly. Notations: before (\textbf{b}), after (\textbf{a}), includes (\textbf{i}), is included (\textbf{ii}), simultaneously (\textbf{s}), vague (\textbf{v}), and $\bar{r}$ represents the reverse relation of $r$.}
\label{tab:trans}
\end{table}

\textbf{Enforcing Linguistic Rules.}
Besides the transitivity constraints represented by $\mc{Y}_1$ above, we also propose to enforce prior knowledge to further constrain the search space for $Y$.
Specifically, linguistic insight has resulted in rules for predicting the temporal relations with special syntactic or semantic patterns, as was done in CAEVO (a state-of-the-art method).
Since these rule predictions often have high-precision, it is worthwhile incorporating them in global reasoning methods as well.

In the CCM framework, these rules can be represented as hard constraints in the search space for $Y$. Specifically,
\begin{equation}\small
\mc{Y}_2 = \left\{Y_j=rule(j), \forall j\in \mc{J}^{(rule)}\right\},
\label{eq:constraint Y2}
\end{equation}
where $\mc{J}^{(rule)}\subseteq \mc{MM}$ is the set of pairs that can be determined by linguistic rules, and $rule(j)\in\mc{R}_T$ is the corresponding decision for pair $j$ according to these rules.
In this work, we used the same set of rules designed by CAEVO for fair comparison.

\subsection{Full Model with Causal Relations}

Now we have presented the joint inference framework for temporal relations in Eq.~\eqref{eq:temp formulation}.
It is easier to explain our complete TCR framework on top of it.
Let $W$ be the vectorization of all causal relations and add the scores from the scoring function for causality $s^c(\cdot)$ to Eq.~\eqref{eq:temp formulation}. Specifically, the full inference formulation is now:
\begingroup\makeatletter\def\f@size{10}\check@mathfonts
\def\maketag@@@#1{\hbox{\m@th\large\normalfont#1}}%
\begin{eqnarray}
\hat{Y}, \hat{W} = \argmax{Y\in\mc{Y},W\in\mc{W}_Y}{}\sum_{i\in\mc{EE}}{}{s^{ee}\{i\mapsto Y_{i}\}}\label{eq:joint formulation}\\
+\sum_{j\in\mc{ET}}{s^{et}\{j\mapsto Y_{j}\}}
+\sum_{k\in\mc{EE}}{}s^c\{k\mapsto W_{k}\}\nonumber
\end{eqnarray}
\endgroup
where $\mc{W}_Y$ is the search space for $W$. $\mc{W}_Y$ depends on the temporal labels $Y$ in the sense that
\begingroup\makeatletter\def\f@size{10}\check@mathfonts
\def\maketag@@@#1{\hbox{\m@th\large\normalfont#1}}%
\begin{eqnarray}
\mc{W}_Y&=&\{W\in\mc{R}_C^m \lvert \forall i,j\in\mc{E}, \textrm{if}~W_{(i,j)}=c, \label{eq:wy}\\
&&\textrm{then}~W_{(j,i)}=\bar{c},~\textrm{and}~Y_{(i,j)}=b\}\nonumber
\end{eqnarray}
\endgroup
where $m$ is the dimension of $W$ (i.e., the total number of causal pairs), $\mc{R}_C=\{c,\bar{c},null\}$ is the label set for causal relations (i.e., ``causes'', ``caused by'', and ``no relation''), and $W_{(i,j)}$ is the causal label for pair $(i,j)$. The constraint represented by $\mc{W}_Y$ means that if a pair of events $i$ and $j$ are labeled to be ``causes'', then the causal relation between $j$ and $i$ must be ``caused by'', and the temporal relation between $i$ and $j$ must be ``before''.

\ignore{Note that some existing work on causality have argued that causality leads to multiple possibilities in temporal, i.e., \rel{before}, \rel{includes}, and \rel{equal}. We find that in the dataset used here, \rel{before} is the dominate relation as a result of causal relation. More details are provided in the discussion section.}

\subsection{Scoring Functions}
In the above, we have built the joint framework on top of scoring functions $s^{ee}(\cdot)$, $s^{et}(\cdot)$ and $s^c(\cdot)$. 
To get $s^{ee}(\cdot)$ and $s^{et}(\cdot)$, we trained classifiers using the averaged perceptron algorithm \cite{FreundSc98} and the same set of features used in \cite{DoLuRo12,NingFeRo17}, and then used the soft-max scores in those scoring functions. For example, that means
\begingroup\makeatletter\def\f@size{10}\check@mathfonts
\def\maketag@@@#1{\hbox{\m@th\large\normalfont#1}}%
$$
s^{ee}\{i\mapsto r\} = \frac{w_r^T \phi(i)}{\sum_{{r^\prime}\in\mc{R}_T}{w_{r^\prime}^T \phi(i)}},~i\in\mc{EE},~r\in\mc{R}_T,
$$
\endgroup
where $\{w_r\}$ is the learned weight vector for relation $r\in\mc{R}_T$ and $\phi(i)$ is the feature vector for pair $i\in\mc{EE}$.

Given a pair of ordered events, we need $s^c(\cdot)$ to estimate the scores of them being ``causes'' or ``caused by''. Since this scoring function has the same nature as $s^{ee}(\cdot)$, we can reuse the features from $s^{ee}(\cdot)$ and learn an averaged perceptron for $s^c(\cdot)$.
In addition to these existing features, we also use prior statistics retrieved using our temporal system from a large corpus\footnote{\url{https://catalog.ldc.upenn.edu/LDC2008T19}, which is disjoint to the test set used here. \QNfinal{Please refer to \cite{NingWuPeRo18} for more analysis on using this corpus to acquire prior knowledge that aids temporal relation classification.}}, so as to know {\em probabilistically} which event happens before another event.
For example, in Example~\ref{ex:prior knowledge 1}, we have a pair of events, \idxevent{1}{died} and \idxevent{2}{exploded}. The prior knowledge we retrieved from that large corpus is that \event{die} happens before \event{explode} with probability 15\% and happens after \event{explode} with probability 85\%.
We think this prior distribution is correlated with causal directionality, so it was also added as features when training $s^c(\cdot)$.

Note that the scoring functions here are implementation choice. The TCR joint framework is fully extensible to other scoring functions. 

\ignore{
Due to the scarcity of  datasets available with causal relations annotated, we adopted the scoring function proposed in \cite{DoChRo11}, which is an unsupervised method.
Specifically, we used the cause-effect association (CEA) metric proposed in \cite{DoChRo11} as the scoring function for causal predictions. 
The CEA between events are defined as the sum of the association scores between their predicates, between the predicate from one event and the argument(s) from the other, and between their arguments. Each of these associations are based on distributional statistics collected from about 760K documents in English Gigaword (LDC2003T05) and do not require supervision.

We realize that the CEA metric itself is symmetric and as a result, the system predictions of \cite{DoChRo11} do not have directionality, i.e., only causal pairs are extracted out, but which one is the cause and which one is the effect is not known. 
To get the directionality, we simply used the appearance order in text and applied the majority classifier for this binary classification problem. 
More sophisticated classifiers can also be used but it is currently beyond the scope of this work.
}

\subsection{Convert the Joint Inference into an ILP}

Conveniently, the joint inference formulation in Eq.~\eqref{eq:joint formulation} can be rewritten into an ILP and solved using off-the-shelf optimization packages, e.g., \cite{Gurobi}. First, we define indicator variables $y_i^r=\mathbb{I}\{Y_i=r\}$, where $\mathbb{I}\{\cdot\}$ is the indicator function, $\forall i\in\mc{MM}$, $\forall r\in\mc{R}_T$.
Then let $p_{i}^r=s^{ee}\{i\mapsto r\}$ if $i\in\mc{EE}$, or $p_{i}^r=s^{et}\{i\mapsto r\}$ if $i\in\mc{ET}$; similarly, let $w_{j}^r=\mathbb{I}\{W_i=r\}$ be the indicator variables for $W_{j}$ and $q_{j}^r$ be the score for $W_{j}=r\in\mc{R}_C$. Therefore, without constraints $\mc{Y}$ and $\mc{W}_Y$ for now, Eq.~\eqref{eq:joint formulation} can be written as:
\begin{equation*}\small
\begin{aligned}
&\hat{y},\hat{w}=\argmax{}{}\sum_{i\in\mc{EE}\cup\mc{ET}}\sum_{r\in\mc{R}_T} p_{i}^r y_{i}^r+\sum_{j\in\mc{EE}}\sum_{r\in\mc{R}_C} q_{j}^r w_{j}^r\\
&\text{s.t.}\qquad y_{i}^r,w_{j}^r\in\{0,1\}, \sum_{r\in\mc{R}_T}y_{i}^r=\sum_{r\in\mc{R}_C}w_{j}^r=1
\end{aligned}
\end{equation*}
The prior knowledge represented as $\mc{Y}$ and $\mc{W}_Y$ can be conveniently converted into constraints for this optimization problem.
Specifically, $\mc{Y}_1$ has two components, symmetry and transitivity:
\begin{equation*}\small
\begin{aligned}
\mc{Y}_1:\qquad 
&\forall i,j,k\in\mc{M},~y_{i,j}^r=y_{j,i}^{\bar{r}},~\text{(symmetry)}\\
&y_{i,j}^{r_1}+y_{j,k}^{r_2}-\sum_{r_3\in\text{Trans}(r_1,r_2)}{y_{i,k}^{r_3}}\le 1~\text{(transitivity)}
\end{aligned}
\end{equation*}
where $\bar{r}$ is the reverse relation of $r$ (i.e., $\bar{b}=a$, $\bar{i}=ii$, $\bar{s}=s$, and $\bar{v}=v$), and Trans$(r_1,r_2)$ is defined in Table~\ref{tab:trans}. 
As for the transitivity constraints, if both $y_{i,j}^{r_1}$ and $y_{j,k}^{r_2}$ are 1, then the constraint requires at least one of $y_{i,k}^{r_3}, r_3\in\text{Trans}(r_1,r_2)$ to be 1, which means the relation between $i$ and $k$ has to be chosen from Trans$(r_1,r_2)$, which is exactly what $\mc{Y}_1$ is intended to do.

The rules in $\mc{Y}_2$ is written as
\begin{equation*}\small
\mc{Y}_2:~ 
y_{j}^r = \mathbb{I}_{\{rule{(j)}=r\}}, \forall j\in\mc{J}^{(rule)}~\text{(linguistic rules)}
\end{equation*}
where $rule{(j)}$  and $\mc{J}^{(rule)}$ have been defined in Eq.~\eqref{eq:constraint Y2}. Converting the $\mathcal{TT}$ constraints, i.e., $\mc{Y}_0$, into constraints is as straightforward as $\mathcal{Y}_2$, so we omit it due to limited space.

Last, converting the constraints $\mc{W}_Y$ defined in Eq.~\eqref{eq:wy} can be done as following:
\begingroup\makeatletter\def\f@size{10}\check@mathfonts
\def\maketag@@@#1{\hbox{\m@th\large\normalfont#1}}%
\begin{equation*}
\mc{W}_Y:~ 
 w_{i,j}^c=w_{j,i}^{\bar{c}}\le y_{i,j}^b,~\forall i,j\in\mc{E}.
\end{equation*}
\endgroup
The equality part, $w_{i,j}^c=w_{j,i}^{\bar{c}}$ represents the symmetry constraint of causal relations;
the inequality part, $w_{i,j}^c\le y_{i,j}^b$ represents that if event $i$ causes event $j$, then $i$ must be before $j$.
\ignore{
This constraint is the bridge between temporal relations and causal relations, based on which we expect that both performances can be improved.
}

\section{Experiments}
In this section, we first show on TimeBank-Dense (TB-Dense) \cite{cassidy2014annotation}, that the proposed framework improves temporal relation identification.
We then explain how our new dataset with both temporal and causal relations was collected, based on which the proposed method improves for both relations.
\begin{table}
	\centering
	\small
	\begin{tabular}{c|c|c|c|c}
		\hline
		\#	&	System	&	P	&	R	&	F$_1$\\
		\hline\hline
		\multicolumn{5}{c}{Ablation Study}\\
		\hline
		1 	&Baseline	&	39.1	&	56.8	&	46.3\\
		2	&+Transitivity$^\dagger$&	42.9	&54.9	&48.2\\
		3	&+$\mc{ET}$	&	44.3	&54.8	&49.0\\
		4	&+Rules&	45.4	&58.7	&51.2\\
		5	&+Causal	&	\textbf{45.8} 	&\textbf{60.5} 	&\textbf{52.1}\\
		\hline\hline
		\multicolumn{5}{c}{Existing Systems$^\ddagger$}\\
		\hline
		6	&ClearTK	&	53.0	&26.4	&35.2\\
		7	&CAEVO		&	\textbf{56.0}	&41.6	&47.8\\
		8	&Ning et al. (2017)& 	47.1& \textbf{53.3}	&\textbf{50.0}\\
		\hline
	\end{tabular}
	
	\begin{tablenotes}
		\small
		\item $^\dagger$This is technically the same with \citet{DoLuRo12}, or \citet{NingFeRo17} without its structured learning component.
		\item $^\ddagger$We added gold $\mc{TT}$ to both gold and system prediction. Without this, Systems 6-8 had F$_1$=28.7, 45.7, and 48.5, respectively, same with the reported values in \citet{NingFeRo17}.
	\end{tablenotes}
	
	\caption{{\bf Ablation study of the proposed system in terms of the standard temporal awareness metric}. The baseline system is to make inference locally for each event pair without looking at the decisions from others. The ``+" signs on lines~2-5 refer to adding a new source of information on top of its preceding system, with which the inference has to be global and done via ILP. All systems are significantly different to its preceding one with p$<$0.05 (McNemar's test).}
	\label{tab:ablation}
\end{table}

\subsection{Temporal Performance on TB-Dense}

Multiple datasets with temporal annotations are available thanks to the  TempEval (TE) workshops \cite{verhagen2007semeval,verhagen2010semeval,uzzaman2013TE3}.
The dataset we used here to demonstrate our improved temporal component was TB-Dense, which was annotated on top of 36 documents out of the classic TimeBank dataset \cite{pustejovsky2003timebank}.
The main purpose of TB-Dense was to alleviate the known issue of sparse annotations in the evaluation dataset provided with TE3 \cite{uzzaman2013TE3}, as pointed out in many previous work \cite{chambers2013navytime,cassidy2014annotation,\CAEVO,NingFeRo17}.
Annotators of TB-Dense were forced to look at each pair of events or timexes within the same sentence or contiguous sentences, so that much fewer links were missed.
Since causal link annotation is not available on TB-Dense, we only show our improvement in terms of temporal performance on TB-Dense. 
\QN{The standard train/dev/test split of TB-Dense was used and parameters were tuned to optimize the F$_1$ performance on dev.
Gold events and time expressions were also used as in existing systems.}

The contributions of each proposed information sources are analyzed in the ablation study shown in Table~\ref{tab:ablation}, where we can see the $F_1$ score was improved step-by-step as new sources of information were added.
Recall that $\mc{Y}_1$ represents transitivity constraints, $\mc{ET}$ represents taking event-timex pairs into consideration, and $\mc{Y}_2$ represents rules from CAEVO \cite{\CAEVO}.
System 1 is the baseline we are comparing to, which is a local method predicting temporal relations one at a time.
System 2 only applied $\mc{Y}_1$ via ILP on top of all $\mc{EE}$ pairs by removing the 2nd term in Eq.~\eqref{eq:temp formulation}; for fair comparison with System 1, we added the same $\mc{ET}$ predictions from System 1. System 3 incorporated $\mc{ET}$ into the ILP and mainly contributed to an increase in precision (from 42.9 to 44.3); we think that there could be more gain if more time expressions existed in the testset. 
With the help of additional high-precision rules ($\mc{Y}_2$), the temporal performance can further be improved, as shown by System 4.
Finally, using the causal extraction obtained via \cite{DoChRo11} in the joint framework, the proposed method achieved the best precision, recall, and $F_1$ scores in our ablation study (Systems 1-5).
According to the McNemar's test \cite{everitt1992analysis,dietterich1998approximate}, all Systems 2-5 were significantly different to its preceding system with p$<$0.05.

The second part of Table~\ref{tab:ablation} compares several state-of-the-art systems on the same test set.
ClearTK \cite{\ClearTK} was the top performing system in TE3 in temporal relation extraction. Since it was designed for TE3 (not TB-Dense), it expectedly achieved a moderate recall on the test set of TB-Dense.
CAEVO \cite{\CAEVO} and \citet{NingFeRo17} were more recent methods and achieved better scores on TB-Dense.
Compared with these state-of-the-art methods, the proposed joint system (System 5) achieved the best F$_1$ score with a major improvement in recall.
We think the low precision compared to System 8 is due to the lack of structured learning, and the low precision compared to System 7 is propagated from the baseline (System~1), which was tuned to maximize its F$_1$ score.
However, the effectiveness of the proposed information sources is already justified in Systems~1-5.

\subsection{Joint Performance on Our New Dataset}
\subsubsection{Data Preparation}
\label{subsubsec:data}
TB-Dense only has temporal relation annotations, so in the evaluations above, we only evaluated our temporal performance.
One existing dataset with both temporal and causal annotations available is the Causal-TimeBank dataset (Causal-TB) \cite{causaltb}. However, Causal-TB is sparse in temporal annotations and is even sparser in causal annotations:
In Table~\ref{tab:stats}, we can see that with four times more documents, Causal-TB still has fewer temporal relations (denoted as T-Links therein), compared to TB-Dense; as for causal relations (C-Links), it has less than two causal relations in each document on average.
\QNfinal{Note that the T-Link sparsity of Causal-TB originates from TimeBank, which is known to have missing links \cite{cassidy2014annotation, NingFeRo17}. The C-Link sparsity was a design choice of Causal-TB in which C-Links were annotated based on only explicit causal markers (e.g., ``A happened {\em because} of B'').}

Another dataset with parallel annotations is CaTeRs \cite{CaTeRs}, which was primarily designed for the Story Cloze Test \cite{storycloze} based on common sense stories. It is different to the newswire domain that we are working on.
Therefore, we decided to augment the EventCausality dataset provided in \citet{DoChRo11} with a modified version of the dense temporal annotation scheme proposed in \citet{cassidy2014annotation} and use this new dataset to showcase the proposed joint approach.

\begin{table}
	\centering
	\small
	\begin{tabular}{c|c|c|c|c}
		\hline
		&Doc&Event&T-Link&C-Link\\
		\hline
		TB-Dense&36&1.6k &5.7k& -\\
		EventCausality&25&0.8k&-&580\\
		Causal-TB&183&6.8k &5.1k &318 \\
		New Dataset &25&1.3k&3.4k&172\\	
		\hline
	\end{tabular}
	
	\caption{\textbf{Statistics of our new dataset with both temporal and causal relations annotated}, compared with existing datasets. T-Link: Temporal relation. C-Link: Causal relation. The new dataset is much denser than Causal-TB  in both T-Links and C-Links.}
	\label{tab:stats}
\end{table}

The EventCausality dataset provides relatively dense causal annotations on 25 newswire articles collected from CNN in 2010. As shown in Table~\ref{tab:stats}, it has more than 20 C-Links annotated per document on average (10 times denser than Causal-TB).
However, one issue is that its notion for events is slightly different to that in the temporal relation extraction regime. To construct parallel annotations of both temporal and causal relations, we preprocessed all the articles in EventCausality using ClearTK to extract events and then manually removed some obvious errors in them.
To annotate temporal relations among these events, \QN{we adopted the annotation scheme from TB-Dense given its success in mitigating the issue of missing annotations with the following modifications.
First, we used a crowdsourcing platform, CrowdFlower, to collect temporal relation annotations.
For each decision of temporal relation, we asked 5 workers to annotate and chose the majority label as our final annotation.
Second, we discovered that comparisons involving ending points of events tend to be ambiguous and suffer from low inter-annotator agreement (IAA), so we asked the annotators to label relations based on the starting points of each event.
This simplification does not change the nature of temporal relation extraction but leads to better annotation quality. For more details about this data collection scheme, please refer to \cite{NingWuRo18} for more details.
}

\begin{table}[htbp!]
	\centering
	\small
	\begin{tabular}{ l|c|c|c|c } 
		\hline\hline
		\multirow{2}{*}{}&\multicolumn{3}{c|}{Temporal}&Causal\\\cline{2-5}&P&R&F$_1$&Accuracy\\
		\hline
		1. Temporal Only&67.2&72.3&69.7&-\\
		2. Causal Only&-&-&-&70.5\\
		3. Joint System&\textbf{68.6}&\textbf{73.8}&\textbf{71.1}&\textbf{77.3}\\
		\hline
		\multicolumn{5}{c}{Enforcing Gold Relations in Joint System}\\
		\hline
		4. Gold Temporal	&	100	&	100	&	100	&	\textit{91.9}\\
		5. Gold Causal		&	\textit{69.3}&	\textit{74.4}&	\textit{71.8}&	100\\
		\hline\hline
	\end{tabular}
	
	\caption{\textbf{Comparison between the proposed method and existing ones, in terms of both temporal and causal performances}.  See Sec.~\ref{subsubsec:data} for description of our new dataset. Per the McNemar's test, the joint system is significantly better than both baselines with p$<$0.05. Lines~4-5 provide the best possible performance the joint system could achieve if gold temporal/causal relations were given.}
	\label{tab:joint}
\end{table}
\subsubsection{Results}
Result on our new dataset jointly annotated with both temporal and causal relations  is shown in Table~\ref{tab:joint}.
We split the new dataset into 20 documents for training and 5 documents for testing. In the training phase, the training parameters were tuned via 5-fold cross validation on the training set.

Table~\ref{tab:joint} demonstrates the improvement of the joint framework over individual components.
The ``temporal only'' baseline is the improved temporal extraction system for which the joint inference with causal links has NOT been applied. 
The ``causal only'' baseline is to use $s^{c}(\cdot)$ alone for the prediction of each pair. That is, for a pair $i$, if $s^{c}\{i\mapsto \text{causes}\}>s^{c}\{i\mapsto \text{caused by}\}$, we then assign ``causes'' to pair $i$; otherwise, we assign ``caused by'' to pair $i$.
\QN{Note that the ``causal accuracy'' column in Table~\ref{tab:joint} was evaluated only on gold causal pairs.}

\QN{In the proposed joint system, the temporal and causal scores were added up for all event pairs.}
The temporal performance got strictly better in precision, recall, and F$_1$, and the causal performance also got improved by a large margin from 70.5\% to 77.3\%, indicating that temporal signals and causal signals are helpful to each other.
According to the McNemar's test, both improvements are significant with p$<$0.05.

The second part of Table~\ref{tab:joint} shows that if gold relations were used, how well each component would possibly perform.
Technically, these gold temporal/causal relations were enforced via adding extra constraints to ILP in Eq.~\eqref{eq:joint formulation} (imagine these gold relations as a special rule, and convert them into constraints like what we did in Eq.~\eqref{eq:constraint Y2}).
When using gold temporal relations, causal accuracy went up to 91.9\%. That is, 91.9\% of the C-Links satisfied the assumption that the cause is temporally before the effect. 
First, this number is much higher than the 77.3\% on line~3, so there is still room for improvement. 
Second, it means in this dataset, there were 8.1\% of the C-Links in which the cause is temporally {\em after} its effect. We will discuss this seemingly counter-intuitive phenomenon in the Discussion section.
When gold causal relations were used (line~5), the temporal performance was slightly better than line~3 in terms of both precision and recall.
The small difference means that the temporal performance on line~3 was already very close to its best.
Compared with the first line, we can see that gold causal relations led to approximately 2\% improvement in precision and recall in temporal performance, which is a reasonable margin given the fact that C-Links are often much sparser than T-Links in practice.

Note that the temporal performance in Table~\ref{tab:joint} is consistently better than those in Table~\ref{tab:ablation} because of the higher IAA in the new dataset.
However, the improvement brought by joint reasoning with causal relations is the same, which further confirms the capability of the proposed approach.

\section{Discussion}
We have consistently observed that on the TB-Dense dataset, if automatically tuned to optimize its F$_1$ score, a system is very likely to have low precisions and high recall (e.g., Table~\ref{tab:ablation}).
We notice that our system often predicts non-vague relations when the TB-Dense gold is “vague”, resulting in lower precision.
However, on our new dataset, the same algorithm can achieve a more balanced precision and recall.
This is an interesting phenomenon, possibly due to the annotation scheme difference which needs further investigation.

The temporal improvements in both Table~\ref{tab:ablation} and Table~\ref{tab:joint} are relatively small (although statistically significant). This is actually not surprising because C-Links are much fewer than T-Links in newswires which focus more on the temporal development of stories.
As a result, many T-Links are not accompanied with C-Links and the improvements are diluted.
But for those event pairs having both T-Links and C-Links, the proposed joint framework is an important scheme to synthesize both signals and improve both. The comparison between Line~5 and Line~3 in Table~\ref{tab:joint} is a showcase of the effectiveness.
\QNfinal{We think that a deeper reason for the improvement achieved via a joint framework is that causality often encodes human’s prior knowledge as global information (e.g., ``death'' is {\em caused by} ``explosion'' rather than {\em causes} ``explosion'', regardless of the local context), while temporality often focuses more on the local context. From this standpoint, temporal information and causal information are complementary and helpful to each other.}

When doing error analysis for the fourth line of Table~\ref{tab:joint}, we noticed some examples that break the commonly accepted temporal precedence assumption.
It turns out that they are not annotation mistakes:
In Example~\ref{ex:assumption fail}, \idxevent{8}{finished} is obviously \rel{before} \idxevent{9}{closed}, but \event{e9} is a cause of \event{e8} since if the market did not close, the shares would not finish.
In the other sentence of Example~\ref{ex:assumption fail}, she prepares \rel{before} hosting her show, but \idxevent{11}{host} is the cause of \idxevent{10}{prepares} since if not for hosting, no preparation would be needed. 
In both cases, the cause is temporally after the effect because people are inclined to make projections for the future and change their behaviors before the future comes.
The proposed system is currently unable to handle these examples and we believe that a better definition of what can be considered as events is needed, as part of further investigating how causality is expressed in natural language. 

\begin{table}
	\centering\small
	\begin{tabular}{|p{7cm}|}
		\hline
		\textbf{Ex~\addexctr\label{ex:assumption fail}: Cause happened after effect.}\\
		\hline
		The shares fell to a record low of \textyen 60 and  \event{(e8:finished)} at \textyen 67 before the market \event{(e9:closed)} for the New Year holidays.\\
		\hline
		As she \event{(e10:prepares)} to \event{(e11:host)} her first show, Crowley writes on what viewers should expect.\\
		\hline
	\end{tabular}
\end{table}

\QNfinal{
Finally, the constraints connecting causal relations to temporal relations are designed in this paper as ``if A is the cause of B, then A must be \rel{before} B". People have suggested other possibilities that involve the \rel{includes} and \rel{simultaneously} relations. While these other relations are simply different interpretations of temporal precedence (and can be easily incorporated in our framework), we find that they rarely happen in our dataset.
}

\section{Conclusion}
We presented a novel joint framework, \textbf{T}emporal and \textbf{C}ausal \textbf{R}easoning (TCR), using CCMs and ILP to the extraction problem of temporal and causal relations between events.
To show the benefit of TCR, we have developed a new dataset that jointly annotates temporal and causal annotations, and then exhibited that TCR can improve both temporal and causal components.
We hope that this notable improvement can foster more interest in jointly studying multiple aspects of events (e.g., event sequencing, coreference, parent-child relations) towards the goal of understanding events in natural language.

\section*{Acknowledgements}
\QNfinal{We thank all the reviewers for providing insightful comments and critiques. This research is supported in part by a grant from the Allen Institute for Artificial Intelligence (allenai.org); the IBM-ILLINOIS Center for Cognitive Computing Systems Research (C3SR) - a research collaboration as part of the IBM AI Horizons Network; by DARPA under agreement number FA8750-13-2-0008; and by the Army Research Laboratory (ARL) under agreement W911NF-09-2-0053 (the ARL Network Science CTA).}

\QNfinal{The U.S. Government is authorized to reproduce and distribute reprints for Governmental purposes notwithstanding any copyright notation thereon. 
The views and conclusions contained herein are those of the authors and should not be interpreted as necessarily representing the official policies or endorsements, either expressed or implied, of DARPA, of the Army Research Laboratory or the U.S. Government.
Any opinions, findings, conclusions or recommendations are those of the authors and do not necessarily reflect the view of the ARL.}

\bibliography{acl2018,emnlp2017,cited-long,ccg-long}
\bibliographystyle{acl_natbib.bst}

\end{document}